\documentclass{article}
\usepackage{flushend} 
\usepackage{balance}

\usepackage{array,multirow,graphicx}
\usepackage{microtype}
\usepackage{graphicx}
\usepackage{subfigure}
\usepackage{booktabs} 

\usepackage{hyperref}

\usepackage[accepted]{sty}
\usepackage{graphicx}
\usepackage{pythonhighlight}

\usepackage{times}
\usepackage{helvet}
\usepackage{courier}
\usepackage{subfigure} 
\usepackage{cancel}
\RequirePackage{amsthm,amsmath}
\usepackage{graphicx, amssymb, mathrsfs,amsfonts,url, bm}

\usepackage{dsfont}
\newcommand{\indicator}{\mathds{1}}
\newcommand{\real}{\mathbb{R}}

\newcommand{\gap}{\,\,\,\,\,\,\,\,}

\newcommand{\btheta}{\boldsymbol\theta}

\newcommand{\blambda}{\boldsymbol\lambda}

\newcommand{\ba}{\bm{a}}
\newcommand{\bA}{\bm{A}}

\newcommand{\bO}{\bm{O}}

\newcommand{\bR}{\bm{R}}

\newcommand{\bw}{\bm{w}}
\newcommand{\bW}{\bm{W}}
\newcommand{\bx}{\bm{x}}

\newcommand{\bz}{\bm{z}}
\newcommand{\bZ}{\bm{Z}}

\newcommand{\exponential}{\mathcal{E}}

\newcommand{\inversegammadist}{\mathcal{G}^{-1}}
\newcommand{\rectifieddist}{\mathcal{RN}}
\newcommand{\normal}{\mathcal{N}}
\newcommand{\truncatednormal}{\mathcal{TN}}

\usepackage[font=small,skip=2pt]{caption}
 
\newcommand{\titlethis}{
Robust Bayesian Nonnegative Matrix Factorization with Implicit Regularizers
}
\icmltitlerunning{\titlethis}

\begin{document}

\twocolumn[
\icmltitle{\titlethis}

\begin{icmlauthorlist}
\icmlauthor{Jun Lu}{te}
\icmlauthor{\gap \gap Christine P. Chai}{t2}
\end{icmlauthorlist}

\icmlaffiliation{te}{Jun Lu, \texttt{jun.lu.locky@gmail.com}.}
\icmlaffiliation{t2}{Christine P. Chai, \texttt{cpchai21@gmail.com}. Microsoft Corporation, Redmond WA 90852 USA. Disclaimer: The opinions and views expressed in this manuscript are those of the author and do not necessarily state or reflect those of Microsoft. Copyright 2022 by the author(s)/owner(s). August 22nd, 2022}
\icmlcorrespondingauthor{Jun Lu}{\texttt{jun.lu.locky@gmail.com}}



\vskip 0.3in
]



\printAffiliationsAndNotice{}  

\begin{abstract}

We introduce a probabilistic model with implicit norm regularization for learning nonnegative matrix factorization (NMF) that is commonly used for predicting missing values and finding hidden patterns in the data, in which the matrix factors are latent variables associated with each data dimension. The nonnegativity constraint for the latent factors is handled by choosing priors with support on the nonnegative subspace, e.g., exponential density or distribution based on exponential function. Bayesian inference procedure based on Gibbs sampling is employed. We evaluate the model on several real-world datasets including Genomics of Drug Sensitivity in Cancer (GDSC $IC_{50}$) and Gene body methylation with different sizes and dimensions, and show that the proposed Bayesian NMF GL$_2^2$ and GL$_{2,\infty}^2$ models lead to robust predictions for different data values and avoid overfitting compared with competitive Bayesian NMF approaches.

\end{abstract}

\section{Introduction}


Over the decades, low-rank matrix approximation methods provide a simple and effective approach to collaborative filtering for modeling user preferences \citep{marlin2003modeling, lim2007variational, mnih2007probabilistic, chen2009collaborative, gillis2020nonnegative, lu2022flexible}. The idea behind such models is that preferences of a user are determined by a small number of unobserved factors \citep{salakhutdinov2008bayesian}. 
The Netflix competition winners, \citet{koren2009matrix}, also employed nonnegative matrix factorization (NMF) in collaborative filtering to build a highly effective recommendation system.
Nowadays nonnegative matrix factorization (NMF) models have remained popular, since the constraint of nonnegativity makes the decompositional parts more interpretable \citep{wang2015multidimensional, song2019improved}.


The goal of nonnegative matrix factorization (NMF) is to find a low rank representation of nonnegative data matrix  as the product of two nonnegative matrices.
Methods for factoring nonnegative matrices fall into two categories -- non-probabilistic and probabilistic. Non-probabilistic methods typically use multiplicative updates for matrix factorization \citep{comon2009tensor, lu2022matrix}. Probabilistic methods mean the factorization is done by maximum-a-posteriori (MAP) or Bayesian inference \citep{mnih2007probabilistic, schmidt2009probabilistic, brouwer2017prior}.
Non-probabilistic solutions give a single point estimate that can easily lead to overfitting. 
For example, the two algorithms, one minimizing least squares error and the other minimizing the Kullback-Leibler divergence, proposed by \citet{lee1999learning, lee2000algorithms} are not robust to sparse data \citep{brouwer2017prior, lu2022flexible}.
Therefore, probabilistic methods are favorable because they can quantify the model order and account for parameter uncertainties.
The MAP estimates maximize the log-posterior over the parameters to train the model. However, the posterior distribution over the factors is intractable, so it is easy to fall into ad-hoc combinations of the parameters \citep{hofmann1999probabilistic, marlin2003modeling, salakhutdinov2008bayesian}.
While we can apply various prior choices to reduce overfitting, Bayesian inference overcomes this issues by locating a full distribution over the nonnegative spaces.

In light of this, our attention is drawn to Bayesian approach for nonnegative matrix factorization.
Given the  matrix $\bA$, the nonnegative factorization can be represented as $\bA=\bW\bZ+\bR\in \real_+^{M\times N}$, where the data matrix is approximately factorized into an $M\times K$ nonnegative matrix $\bW$ and a $K\times N$ nonnegative matrix $\bZ$; the residuals are captured by matrix $\bR\in \real^{M\times N}$ (having both positive and nonnegative entries). 
The first matrix $\bW$ contains low-rank column basis of the data matrix $\bA$ in columns, while $\bZ$ comprises row basis of $\bA$ in rows.
It is also possible that the data matrix $\bA$ is sparse and incomplete, and the indices of observed entries can be represented by a mask matrix $M\times N$ matrix $\bO$ that contains values of 0 and 1 to indicate the observation of each entry. 
The missing entries can be easily handled in Bayesian NMF inference by excluding the missing elements in the likelihood term.
To be more concrete, in the Netflix user preference context, the factorization means that the $M\times N$ preference matrix of rating that $M$ users assign to $N$ movies is modeled by the product of an $M\times K$ user feature matrix $\bW$ and a $K\times N$ movie feature matrix $\bZ$ \citep{srebro2003weighted, salakhutdinov2008bayesian}. 

Project data vector $\ba_n$ ($n$-th column of $\bA$) to a smaller dimension $\bz_n \in \real^K$  with $K<M$,
such that the \textit{reconstruction error} measured by mean squared error (MSE) is minimized (assume $K$ is known):
\begin{equation}\label{equation:als-per-example-loss}
	\mathop{\min}_{\bW,\bZ} \,\, \sum_{n=1}^N \sum_{m=1}^{M} \left(a_{mn} - \bw_m^\top\bz_n\right)^2 \cdot o_{mn},
\end{equation}
where $\bW=[\bw_1^\top; \bw_2^\top; \ldots; \bw_M^\top]\in \real_+^{M\times K}$ and $\bZ=[\bz_1, \bz_2, \ldots, \bz_N] \in \real_+^{K\times N}$ contain $\bw_m$'s and $\bz_n$'s as \textbf{rows and columns} respectively, and $a_{mn}, o_{mn}$ are the $(m,n)$-th entries of data matrix $\bA$ and mask matrix $\bO$ respectively. The term in Eq.~\eqref{equation:als-per-example-loss} is also known as the \textit{Frobenius norm}. And it can be equivalently written as
\begin{equation}\label{equation:loss_nmf_general}
\begin{aligned}
L(\bW,\bZ) &= \sum_{n=1}^N \sum_{m=1}^{M}\left(a_{mn} - \bw_m^\top\bz_n\right)^2 \cdot  o_{mn} \\
&= ||(\bW\bZ-\bA)\odot \bO||^2 ,
\end{aligned}
\end{equation}
where $\odot$ represents the \textit{Hadamard product (element-wise product)} between matrices.

We approach the nonnegative constraint by considering the NMF model as a latent factor model and we describe a fully specified graphical model for the problem and employ Bayesian learning methods to infer the latent factors. In this sense, explicit nonnegativity constraints are not required on the latent factors, since this is naturally taken care of by the appropriate choice of prior distribution, e.g., exponential density, half-normal density, truncated-normal density, or rectified-normal prior.

The main contribution of this paper is to propose a novel Bayesian NMF method which
has implicit $L_p$ norm regularization behind the model so that the models are more robust in various data types.
We propose the Bayesian model called \textit{GL$_2^2$} and \textit{GL$_{2,\infty}^2$} NMF algorithms to both increase convergence performance and out-of-sample accuracy. 
While previous works propose somewhat algorithms that also have implicit regularization meaning (e.g., the GL$_1^2$ model in \citet{brouwer2017prior}), an interpretation on the posterior parameters reveals that model is not robust especially when the entries of the observed matrix $\bA$ are large, in which case, the GL$_1^2$ model tends to impose a regularization far to much and the end result lacks predictive ability.
On the other hand, the proposed GL$_2^2$ and GL$_\infty$ models have simple conditional density forms. 
We show that our methods can be successfully applied to the sparse and imbalanced GDSC $IC_{50}$  dataset; and also to the dense and balanced Gene body methylation dataset.
We also show that the proposed GL$_2^2$ and GL$_\infty$ models significantly increase the models' predictive accuracy (out-of-sample performance),
compared with the standard Bayesian NMF models that has implicit interpretation of norm regularization. 


\section{Related Work}

In this section, we review the Gaussian Exponential (GEE) model for computing nonnegative matrix factorization and its implicit regularization meaning behind the model. 

\subsection{Gaussian Exponential (GEE) Model}

\begin{figure}[!ht]
\centering  
\advance\leftskip-8pt
\vspace{-0.35cm} 
\subfigtopskip=2pt 
\subfigbottomskip=6pt 
\subfigcapskip=-5pt 
\subfigure[GEE.]{\label{fig:bmf_gee}
\includegraphics[width=0.52\linewidth]{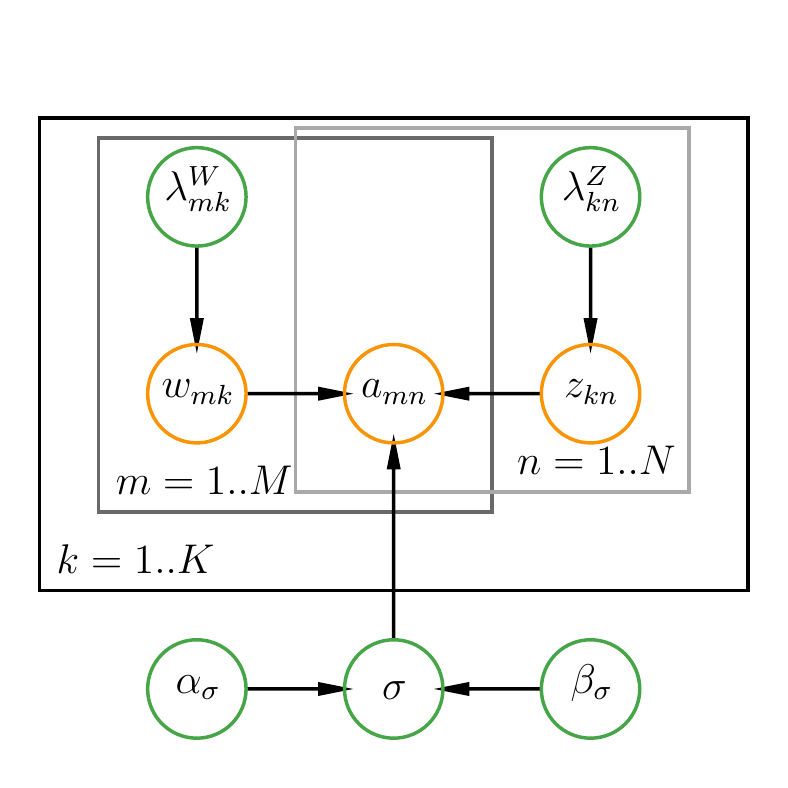}}
\hspace{-1.1em}
\subfigure[GL$_1^2$, GL$_2^2$, GL$_\infty$, GL$_{2,\infty}^2$.]{\label{fig:bmf_gl12}
\includegraphics[width=0.52\linewidth]{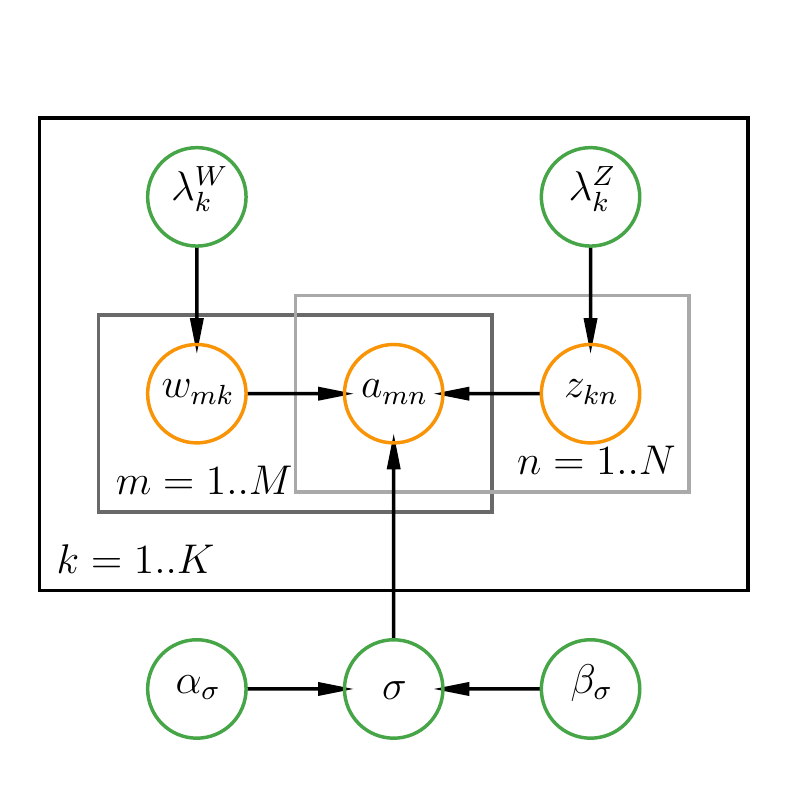}}
\caption{Graphical representation of GL$_1^2$, GL$_2^2$, GL$_\infty$, and GL$_{2,\infty}^2$ models. Orange circles represent observed and latent variables, green circles denote prior variables, and plates represent repeated variables. $\sigma^2$ denotes the variance parameter of normal distribution, and $\alpha_\sigma, \beta_\sigma$ are the parameters of the prior over $\sigma^2$.}
\label{fig:bmf_gl12_gee_nmf_regu}
\end{figure}

We view the data $\bA$ as being produced according to the probabilistic generative process shown in Figure~\ref{fig:bmf_gee}. The observed $(m,n)$-th entry $a_{mn}$ of matrix $\bA$ is modeled using a Gaussian likelihood function with variance $\sigma^2$ and mean given by the latent decomposition $\bw_m^\top\bz_n$ (Eq.~\eqref{equation:als-per-example-loss}):
\begin{equation}\label{equation:grrn_data_entry_likelihood}
p(a_{mn} | \bw_m^\top\bz_n, \sigma^2) = \normal(a_{mn}|\bw_m^\top\bz_n, \sigma^2),
\end{equation}
where $\normal(a_{mn}|\bw_m^\top\bz_n, \sigma^2) =\sqrt{\frac{ 1}{2\pi \sigma^2}}\exp\{-\frac{1}{2\sigma^2} (a_{mn}-\bw_m^\top\bz_n)^2\}$ is a Gaussian distribution with mean $\mu$ and variance $\sigma^2$.

We choose a conjugate prior over the data variance, an inverse-Gamma distribution with shape $\alpha_\sigma$ and scale $\beta_\sigma$, 
\begin{equation}\label{equation:prior_grrn_gamma_on_variance}
	p(\sigma^2 | \alpha_\sigma, \beta_\sigma) = \inversegammadist(\sigma^2 | \alpha_\sigma, \beta_\sigma),
\end{equation}
where $\inversegammadist(\sigma^2 |\alpha_\sigma, \beta_\sigma)= \frac{(\beta_\sigma)^{(\alpha_\sigma)}}{\Gamma(\alpha_\sigma)} (\sigma^2) ^{-\alpha_\sigma-1}$ $\cdot \exp\{-\frac{\beta_\sigma}{\sigma^2 }\}$ $\cdot u(\sigma^2 )$ is an inverse-Gamma distribution, $\Gamma(\cdot)$ is the gamma function, and $u(\sigma^2 )$ is the unit step function that has a value of $1$ when $\sigma^2  > 0$ and $0$ otherwise.

While it can also be equivalently given a conjugate Gamma prior over the precision (inverse of variance) and we shall not repeat the details.

We treat the latent variables $w_{mk}$'s (and $z_{kn}$'s) as random variables. And we need prior densities over these latent variables to express beliefs for their values, e.g., nonnegativity in this context though there are many other constraints (semi-nonnegativity in \citet{ding2008convex}, interpolative prior in \citet{lu2022bayesian, lu2022comparative}, and discreteness in \citet{gopalan2014bayesian, gopalan2015scalable}).
Here we assume further that the latent variables $w_{mk}$'s are independently drawn from an exponential prior
\begin{equation}\label{equation:rn_prior_gee}
w_{mk}\sim \exponential(w_{mk}| \lambda_{mk}^W),
\end{equation}
where $\exponential(w_{mk}|\lambda_{mk}^W) = \lambda_{mk}^W\exp\{-\lambda_{mk}^W \cdot w_{mk}\}u(w_{mk})$ is an exponential distribution.

Similarly, the latent variables $z_{kn}$'s are also drawn from the same exponential prior. This prior serves to enforce the nonnegativity constraint on the components $\bW, \bZ$, and the conditional posterior density is a \textit{truncated-normal distribution}.
In some cases, the two sets of latent variables can be drawn from two different exponential priors (with different $\lambda_{mk}^W$ and $\lambda_{kn}^Z$ parameters for each component; see Figure~\ref{fig:bmf_gee}), e.g., enforcing sparsity in $\bW$ while non-sparsity in $\bZ$. However, this is not the main interest of this paper and we shall not consider this scenario.



\subsection{Priors as Regularization}
Denote the prior parameters as $\btheta$ and follow the Bayes' rule, the posterior is proportional to product of likelihood and prior density:
$
p(\btheta | \bA) \propto p(\bA | \btheta) \cdot p(\btheta) ,
$
such that the log-likelihood follows
$$
\begin{aligned}
&\gap \log p(\btheta | \bA) = \log p(\bA | \btheta) + \log p(\btheta) + C_1  \\ 
&=\log \prod_{m,n=1}^{M,N} \normal \left(a_{mn}| \bw_m^\top\bz_n, \sigma^2 \right) + \log p(\bW,\bZ) + C_2\\
&=-\frac{1}{2\sigma^2} \left(a_{mn} - \bw_m^\top\bz_n \right)^2  + \log p(\bW,\bZ) + C_3,
\end{aligned}
$$
where $C_1,C_2,C_3$ are constants. The ultimate equation is the sum of the negative squared loss of the training fit and a regularization term over the factored components $\bW,\bZ$. The prior distributions of $\bW,\bZ$ then act as a regularization that can prevent the model from overfitting the data and increase the predictive performance. To be more concrete, the regularizers on $\bW$ fall into four categories:
\begin{equation}\label{equation:6norms-in-vanilla-bmf}
\begin{aligned}
L_1 &= \sum_{m=1}^{M} \sum_{k=1}^{K} w_{mk}, \gap    
L_2^{1/2} = \sum_{m=1}^{M} \sqrt{\sum_{k=1}^{K}w_{mk}},\\
L_1^2 &= \sum_{m=1}^{M} \left(\sum_{k=1}^{K} w_{mk}\right)^2,  \gap 
L_2^2 = \sum_{m=1}^{M}{\sum_{k=1}^{K}w_{mk}^2}.
\end{aligned}
\end{equation}
We note that the $L_2^2$ norm 
is equivalent to an independent Gaussian prior (GGG model in \citet{brouwer2017prior}); the $L_1$ norm is equivalent to a Laplace prior (GLL model in \citet{brouwer2017prior}) in real-valued decomposition and is equivalent to an exponential prior (GEE model) in nonnegative matrix factorization.

\subsection{Gaussian $L_1^2$ Prior (GL$_1^2$) Model}
The Gaussian $L_1^2$ prior model follows immediately by replacing the $L_1$ norm with $L_1^2$ in the exponential prior:
\begin{equation}\label{equation:gl12_prior_density}
\begin{aligned}
p(\bW|\lambda_k^W) \propto \exp \big\{  \frac{-\lambda_k^W}{2} \sum_{m=1}^{M} \big(\sum_{k=1}^{K} w_{mk}\big)^2 \big\} u(\bW),\\
\end{aligned}
\end{equation}
where $u(\bW)$ denotes that all entries of $\bW$ are nonnegative. A similar prior is placed over component $\bZ$ (see Figure~\ref{fig:bmf_gl12}).

\begin{table*}[t]
\setlength{\tabcolsep}{5.2pt}
\begin{tabular}{l|l|l|l}
\hline
& Conditional $w_{mk}$& $\widetilde{\mu_{mk}}$ (mean) & $\widetilde{\sigma_{mk}^{2}}$  (variance)    \\ \hline
GEE         & $\truncatednormal(w_{mk} | \widetilde{\mu_{mk}}, \widetilde{\sigma_{mk}^{2}})$ & $\left( -\lambda_{mk}^W\gap \gap\gap \,\,\,\,\,+ \frac{1}{\sigma^2} \sum_{j=1}^{N} z_{kj}\big( a_{mj} - \sum_{i\neq k}^{K}w_{mi}z_{ij}\big)  \right) \widetilde{\sigma_{mk}^{2}}$                                         & $ \frac{\sigma^2}{\sum_{j=1}^{N} z_{kj}^2}$                                    \\ 
\hline
GL$_1^2$    & $\truncatednormal(w_{mk} | \widetilde{\mu_{mk}}, \widetilde{\sigma_{mk}^{2}})$ & $\left( -\lambda_k^W\cdot\textcolor{red}{\sum_{j\neq k}^{K}w_{mj}}+\frac{1}{\sigma^2} \sum_{j=1}^{N} z_{kj}\big( a_{mj} - \sum_{i\neq k}^{K}w_{mi}z_{ij}\big)  \right) \widetilde{\sigma_{mk}^{2}}$ & $\frac{\sigma^2}{\sum_{j=1}^{N} z_{kj}^2 +\textcolor{red}{\sigma^2\lambda_k^W}}$ \\ 
\hline
GL$_2^2$    & $\truncatednormal(w_{mk} | \widetilde{\mu_{mk}}, \widetilde{\sigma_{mk}^{2}})$ & $\left(\gap \gap\gap\gap\gap\,\,\,\, \gap\frac{1}{\sigma^2} \sum_{j=1}^{N} z_{kj}\big( a_{mj} - \sum_{i\neq k}^{K}w_{mi}z_{ij}\big)  \right) \widetilde{\sigma_{mk}^{2}}$                                                           & $\frac{\sigma^2}{\sum_{j=1}^{N} z_{kj}^2 +\textcolor{red}{\sigma^2\lambda_k^W}}$ \\ 
\hline
GL$_\infty$ & $\truncatednormal(w_{mk} | \widetilde{\mu_{mk}}, \widetilde{\sigma_{mk}^{2}})$ & $\left(-\textcolor{red}{\lambda_k^W\cdot \indicator(w_{mk})}
\,\,\,\,\,\,\, +\frac{1}{\sigma^2} \sum_{j=1}^{N} z_{kj}\big( a_{mj} - \sum_{i\neq k}^{K}w_{mi}z_{ij}\big)  \right) \widetilde{\sigma_{mk}^{2}}$       & $\frac{\sigma^2}{\sum_{j=1}^{N} z_{kj}^2 }$                                    \\ 
\hline
GL$_{2,\infty}^2$ & $\truncatednormal(w_{mk} | \widetilde{\mu_{mk}}, \widetilde{\sigma_{mk}^{2}})$ & $\left(-\textcolor{red}{\lambda_k^W\cdot \indicator(w_{mk})}
\,\,\,\,\,\,\, +\frac{1}{\sigma^2} \sum_{j=1}^{N} z_{kj}\big( a_{mj} - \sum_{i\neq k}^{K}w_{mi}z_{ij}\big)  \right) \widetilde{\sigma_{mk}^{2}}$       & $\frac{\sigma^2}{\sum_{j=1}^{N} z_{kj}^2 +
\textcolor{red}{\sigma^2\lambda_k^W} 
}$                                    \\ 
\hline
\end{tabular}
\caption{Posterior conditional densities of $w_{mk}$'s for GEE, GL$_1^2$, GL$_2^2$, GL$_\infty$, and GL$_{2,\infty}^2$ models. 
The difference is highlighted in red.	
The conditional densities of $z_{kn}$'s are similar due to their symmetry to $w_{mk}$'s.
$\truncatednormal(x|\mu,\tau^{-1}) =\frac{\sqrt{\frac{\tau}{2\pi}} \exp\{-\frac{\tau}{2} (x-\mu)^2 \} } 
{1-\Phi(-\mu\sqrt{\tau})} u(x)$
is a truncated-normal (TN) density with zero density below $x=0$ and renormalized to integrate to one. $\mu$ and $\tau$ are known as the ``parent" mean and ``parent" precision. $\Phi(\cdot)$ is the cumulative distribution function of standard normal density $\normal(0,1)$.
}
\label{table:nmf_regularizer_posterior}
\end{table*}

\section{Gaussian $L_p$ Prior Models}
\begin{figure}[h]
\centering  
\vspace{-0.25cm} 
\subfigtopskip=2pt 
\subfigbottomskip=9pt 
\subfigcapskip=0pt 
\includegraphics[width=0.3\textwidth]{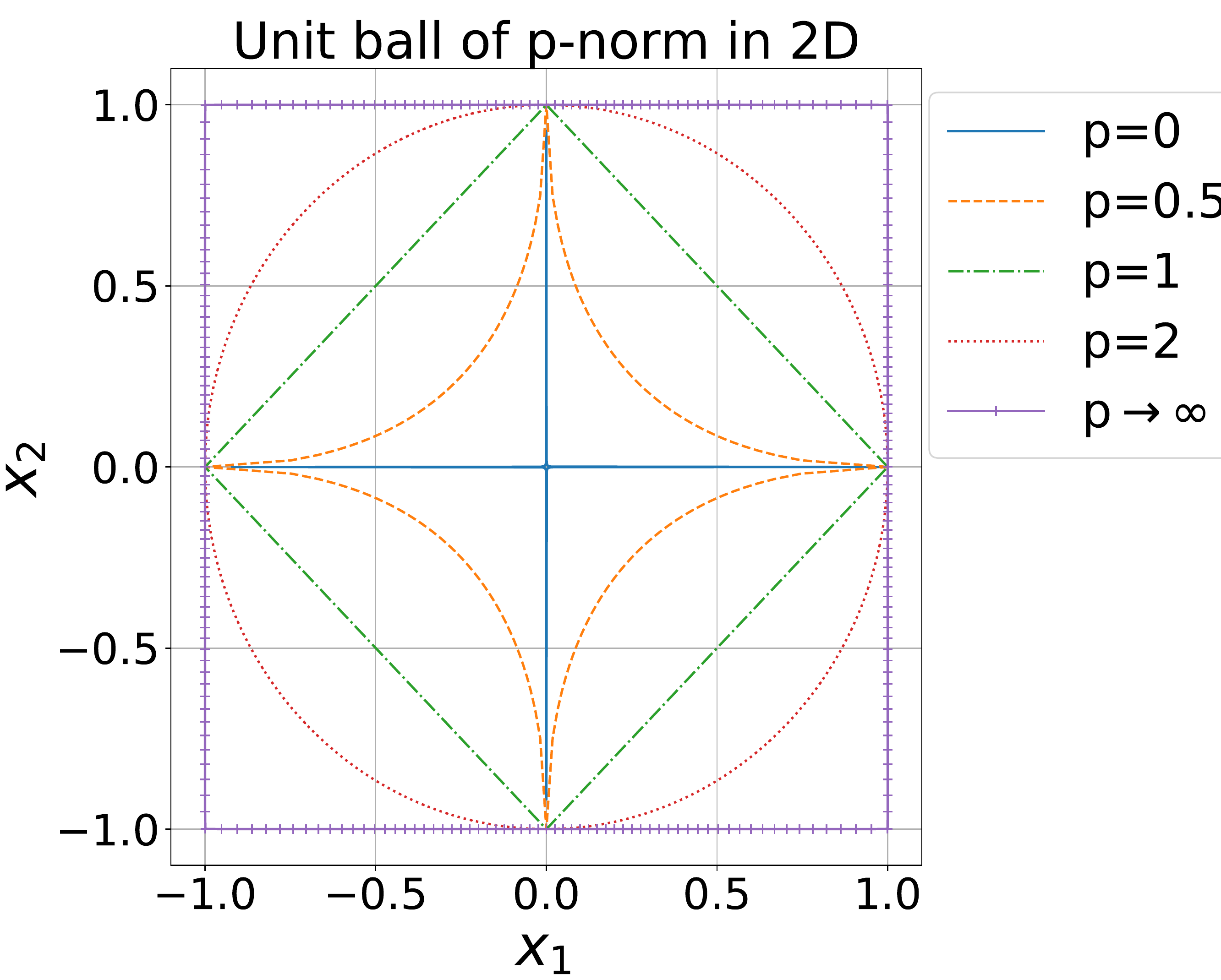}
\caption{Unit ball of $L_p$ norm in 2-dimensional space.}
\label{fig:nmf_reg_pnorm}
\end{figure}

The proposed Gaussian $L_p$ prior models highly rely on the implicit regularization in GL$_1^2$ models. For any vector $\bx\in \real^n$, the $L_p$ norm is given by 
$
L_p(\bx) = \left(\sum_{i=1}^{n} |x_i|^p\right)^{1/p}
$
whose unit ball in 2-dimensional space is shown in Figure~\ref{fig:nmf_reg_pnorm}.

The $L_1$ norm is known to have \textit{sparse constraint}.
We further extend the Bayesian models with $L_2^2$ and $L_\infty$ norms. We will show that the $L_2^2$ and  $L_\infty$ norms in the Bayesian NMF context have this sparse constraint as well.
To be more concrete, the prior for $\bW$ of \textit{GL$_2^2$ model} is given by 
\begin{equation}\label{equation:gp22_prior_density}
p(\bW|\lambda_k^W) \propto\exp \big\{  -\frac{\lambda_k^W}{2} \sum_{m=1}^{M} \big(\sum_{k=1}^{K} w_{mk}^2\big) \big\} u(\bW).
\end{equation}
And the prior for $\bW$ of \textit{GL$_\infty$ model} is given by 
\begin{equation}\label{equation:gpinfty_prior_density}
\begin{aligned}
p(\bW|\lambda_k^W) \propto \exp \big\{  -{\lambda_k^W} \sum_{m=1}^{M} \mathop{\max}_{k} |w_{mk}| \big\} u(\bW).
\end{aligned}
\end{equation}
Note we remove the value of 2 in Eq.~\eqref{equation:gpinfty_prior_density} for the denominator of $\lambda_k^W$ for consistency issue which we will see shortly in the form of the conditional density in Table~\ref{table:nmf_regularizer_posterior} or Eq.~\eqref{equation:gpinfty_poster_wmk1_append} in Appendix~\ref{appendix:glinfity_regu_nmf}.

\subsection{Gibbs Sampler}\label{section:gibbs_nmf_regu}

Here we use Gibbs sampling because it is easier and accurate to sample from the conditional distributions than the joint distribution \citep{hoff2009first}. Alternative methods are variational Bayesian inference or Metropolis-Hastings, but we shall not go into the details \citep{tichy2019bayesian}. We shortly describe the posterior conditional density in this section, and the detailed derivation can be found in Appendix~\ref{appendix:gibbs_NMF_regularizer}.
The conditional density of $\sigma^2$ is an inverse-Gamma distribution by conjugacy,
\begin{equation}\label{equation:posterior_sigma_nmfregu_sigma2}
	\begin{aligned}
		&\gap p(\sigma^2 | \bW, \bZ, \bA)
		= \inversegammadist(\sigma^2 | \widetilde{\alpha_\sigma}, \widetilde{\beta_\sigma}),
	\end{aligned}
\end{equation}
where $\widetilde{\alpha_\sigma} = \frac{MN}{2}+\alpha_\sigma$, 
$\widetilde{\beta_\sigma}=\frac{1}{2} \sum_{i,j=1}^{M,N}(a_{ij}-\bw_i^\top\bz_j)^2+\beta_\sigma$.

The conditional density of $w_{mk}$'s for  GEE, GL$_1^2$, GL$_2^2$, and GL$_\infty$ models are summarized in Table~\ref{table:nmf_regularizer_posterior} where the difference is highlighted in red. A detailed derivation is provided in Appendix~\ref{appendix:gibbs_NMF_regularizer}.
The posterior conditional density of $z_{kn}$'s can be derived in a similar way.
The full procedure is formulated in Algorithm~\ref{alg:nmf_regular_gibbs_sampler}.

\begin{algorithm}[h] 
\caption{Gibbs sampler for GEE, GL$_1^2$, GL$_2^2$, and GL$_\infty$ models (prior on the variance parameter $\sigma^2$). The procedure presented here is for explanatory purposes, and vectorization can expedite the procedure. 
Users need to specify the total number of iterations $T$, the observed matrix $\bA$ of shape $M\times N$, and the latent dimension $K$.
By default, uninformative priors are $\{\lambda_{mk}^W\}=\{\lambda_{kn}^Z\}=0.1$ (GEE); 
$\{\lambda_{k}^W\}=\{\lambda_{k}^Z\}=0.1$ (GL$_1^2$, GL$_2^2$, GL$_\infty$, GL$_{2,\infty}^2$);
$\alpha_\sigma=\beta_\sigma=1$ (inverse-Gamma prior in GEE, GL$_1^2$, GL$_2^2$, GL$_\infty$, GL$_{2,\infty}^2$); 
} 
\label{alg:nmf_regular_gibbs_sampler}  
\begin{algorithmic}[1] 
\FOR{$t=1$ to $T$} 		
\FOR{$k=1$ to $K$} 
\FOR{$m=1$ to $M$}
\STATE Sample $w_{mk}$ from $p(w_{mk} | \cdot )$ from Table~\ref{table:nmf_regularizer_posterior}; 
\ENDFOR
\FOR{$n=1$ to $N$}
\STATE Sample $z_{kn}$ from $p(z_{kn} |\cdot )$ (symmetry of $w_{mk}$); 
\ENDFOR
\ENDFOR
\STATE Sample $\sigma^2$ from $p(\sigma^2 | \bW,\bZ, \bA)$ (Eq.~\eqref{equation:posterior_sigma_nmfregu_sigma2}); 
\ENDFOR
\end{algorithmic} 
\end{algorithm}

\paragraph{Sparse constraint in GEE}
We note that there is a negative term $-\lambda_{mk}^W$ 
in the posterior ``parent" mean parameter $\widetilde{\mu_{mk}}$ for GEE in Table~\ref{table:nmf_regularizer_posterior} that can push the posterior ``parent" mean $\widetilde{\mu_{mk}}$ towards zero or negative values. The draws of $\truncatednormal(w_{mk}| \widetilde{\mu_{mk}}, \widetilde{\sigma_{mk}^2})$ will then be around zero thus imposing sparsity.

\paragraph{Connection between GEE and GL$_1^2$ models}
The second term $\sigma^2\lambda_k^W$ exists in the GL$_1^2$ denominator of the variance $\widetilde{\sigma_{mk}^{2}}$. When all else are held equal, the GL$_1^2$ has smaller variance than GEE, so the distribution of GL$_1^2$ is more clustered in a smaller range. This is actually a stronger constraint/regularizer than  the GEE model. 

Moreover, when $\{\lambda_{mk}^W\}$ in GEE model and $\{\lambda_k^W\}$ in GL$_1^2$ model are equal, the extra term $\sum_{j\neq k}^{K}w_{mj}$ in  GL$_1^2$ model plays an important role in controlling the sparsity of factored components in NMF context. To be more concrete, when the distribution of elements in matrix $\bA$ has a large portion of big values, the extra term $\sum_{j\neq k}^{K}w_{mj}$ will be larger than 1 and thus enforce the posterior ``parent" mean $\widetilde{\mu_{mk}}$ of the truncated-normal density to be a small positive or even a negative value. This in turn constraints the draws of $p(w_{mk}|\cdot)$ to be around zero thus favoring sparsity (see Section~\ref{section:nmf_regulari_experiments} for the experiment on GDSC $IC_{50}$ dataset).
On the contrary, when the entries in matrix $\bA$ are small, this extra term will be smaller than 1, the parameter $\lambda_k^W$ has little impact on the posterior ``parent" mean $\widetilde{\mu_{mk}}$ which will possibly be a large value, and the factored component $\bW$ or $\bZ$ will be dense instead (also see Section~\ref{section:nmf_regulari_experiments} for the experiment on Gene body methylation dataset).

In this sense, the drawback of the GL$_1^2$ model is revealed that it is not consistent and not robust for different types of the matrix $\bA$.
In contrast, the proposed GL$_2^2$ and GL$_{2,\infty}^2$ models are consistent and robust for different matrix types and impose a larger regularization compared with the GEE model such that its predictive performance is better (when the data matrix $\bA$ has large values).

\paragraph{Connection between GEE, GL$_1^2$, and GL$_2^2$ models}
We observe that the posterior ``parent" mean $\widetilde{\mu_{mk}}$ in the GL$_2^2$ model is larger than that in the GEE model since it does not contain the negative term $-\lambda_{mk}^W$. 
While the posterior ``parent" variance is smaller than that in the GEE model, such that the 
conditional density of GL$_2^2$ model is more clustered and it imposes a larger regularization in the sense of data/entry distribution. This can induce sparsity in the context of nonnegative matrix factorization.
 Moreover, the GL$_2^2$ does not have the 
extra term $\sum_{j\neq k}^{K}w_{mj}$ in  GL$_1^2$ model which causes the inconsistency for different types of matrix $\bA$ such that the proposed GL$_2^2$ model is more robust.

\paragraph{Connection between GEE and GL$_\infty$ models}
The posterior ``parent" variance $\widetilde{\sigma^2_{mk}}$ in the GL$_2^2$ model is exactly the same as  that in the GEE model.
Denote $\indicator(w_{mk})$ as the indicator whether $w_{mk}$ is the largest one among $k=1,2,\ldots, K$.
Suppose further the condition $\indicator(w_{mk})$ is satisfied, parameters $\{\lambda_{mk}^W\}$ in GEE model and $\{\lambda_k^W\}$ in GL$_\infty$ model are equal, the ``parent" mean $\widetilde{\mu_{mk}}$ is the same as that in the GEE model as well.
However, when $w_{mk}$ is not the maximum value among $\{w_{m1}, w_{m2}, \ldots, w_{mK}\}$, the ``parent" mean $\widetilde{\mu_{mk}}$ is larger than that in the GEE model since the GL$_\infty$ model excludes this negative term.
The GL$_\infty$ model then has the interpretation that it has a \textit{sparsity constraint} when $w_{mk}$ is the maximum value; and it has a \textit{relatively loose constraint} when $w_{mk}$ is not the maximum value. Overall, the GL$_\infty$ favors a loose regularization compared with the GEE model.

\paragraph{Further extension: GL$_{2,\infty}^2$ model}

The GL$_{2,\infty}^2$ model takes the advantages of both GL$_2^2$ and GL$_\infty$, and the posterior parameters of GL$_{2,\infty}^2$ are shown in Table~\ref{table:nmf_regularizer_posterior}.
The implicit prior of the GL$_{2,\infty}^2$ model can be obtained by 
\begin{equation}\label{equation:gp22infty_prior_density}
\begin{aligned}
&\gap p(\bW|\lambda_k^W) \propto\\
&\exp \big\{  \frac{-\lambda_k^W}{2} \sum_{m=1}^{M} \big(\sum_{k=1}^{K} w_{mk}^2+2\mathop{\max}_{k} |w_{mk}| \big) \big\} u(\bW).
\end{aligned}
\end{equation}

\paragraph{Computational complexity}
The adopted Gibbs sampling methods for GEE, GL$_1^2$, GL$_2^2$, GL$_\infty$, and GL$_{2,\infty}^2$ models have complexity $\mathcal{O}(MNK^2)$, where the most expensive operation is the update on the conditional density of $w_{mk}$'s and $z_{kn}$'s.

\begin{figure*}[htp]
\centering  
\vspace{-0.25cm} 
\subfigtopskip=2pt 
\subfigbottomskip=2pt 
\subfigcapskip=-1pt 
\subfigure[Convergence on the \textbf{GDSC $\boldsymbol{IC_{50}}$} dataset with increasing latent dimension $K$.]{\includegraphics[width=1\textwidth]{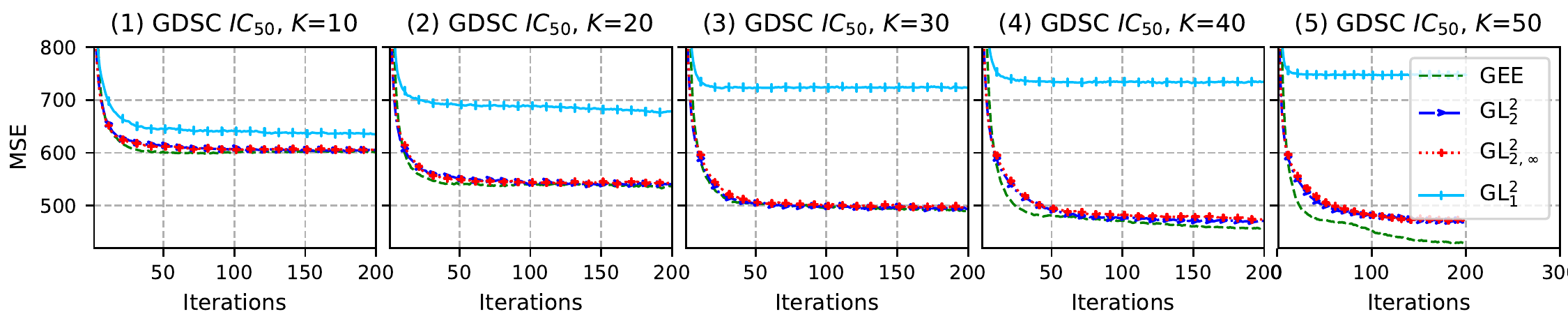} \label{fig:nmf_regularizers_convergences_gdsc}}\vspace{-0.6em}
\subfigure[Data distribution of factored component $\bW$ in the last 20 iterations for \textbf{GDSC $\boldsymbol{IC_{50}}$}. ]{\includegraphics[width=1\textwidth]{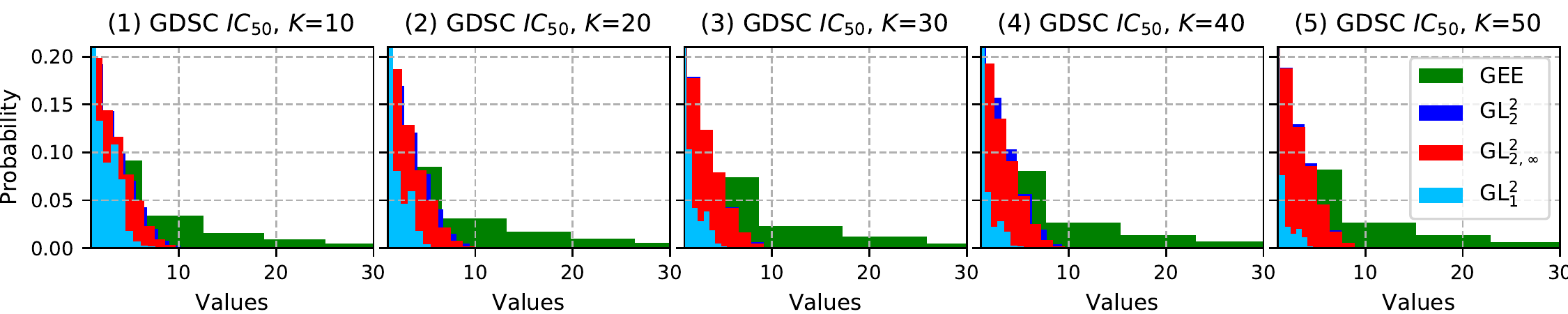} \label{fig:nmf_regularizers_distributions_gdsc}}\vspace{-0.1em}
\caption{Convergence of the models on the 
GDSC $IC_{50}$ (upper) and the distribution of factored $\bW$  (lower), measuring the training data fit (mean squared error). When we increase latent dimension $K$, the GEE and the proposed GL$_2^2$ and GL$_\infty$ algorithms continue to increase the performance; while GL$_1^2$ start to decrease.}
\label{fig:convergences_gdsc_nmf_regularizer}
\end{figure*}

\begin{figure*}[h]
\centering  
\vspace{-0.25cm} 
\subfigtopskip=2pt 
\subfigbottomskip=2pt 
\subfigcapskip=2pt 
\subfigure[Convergence on the \textbf{Gene body methylation} dataset with increasing latent dimension $K$.]{\includegraphics[width=1\textwidth]{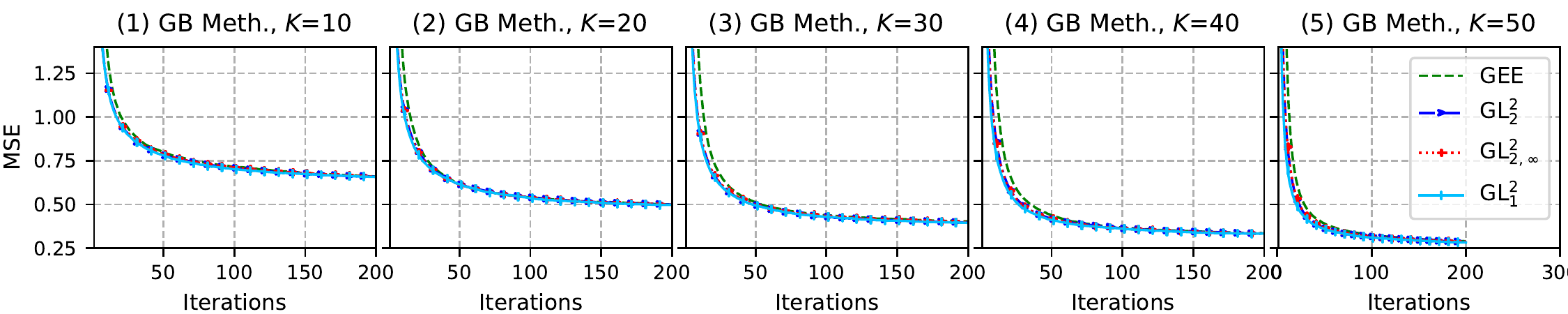} \label{fig:nmf_regularizers_convergences_gene_body}}\vspace{-0.6em}
\subfigure[Data distribution of factored component $\bW$ in the last 20 iterations for \textbf{Gene body methylation}.]{\includegraphics[width=1\textwidth]{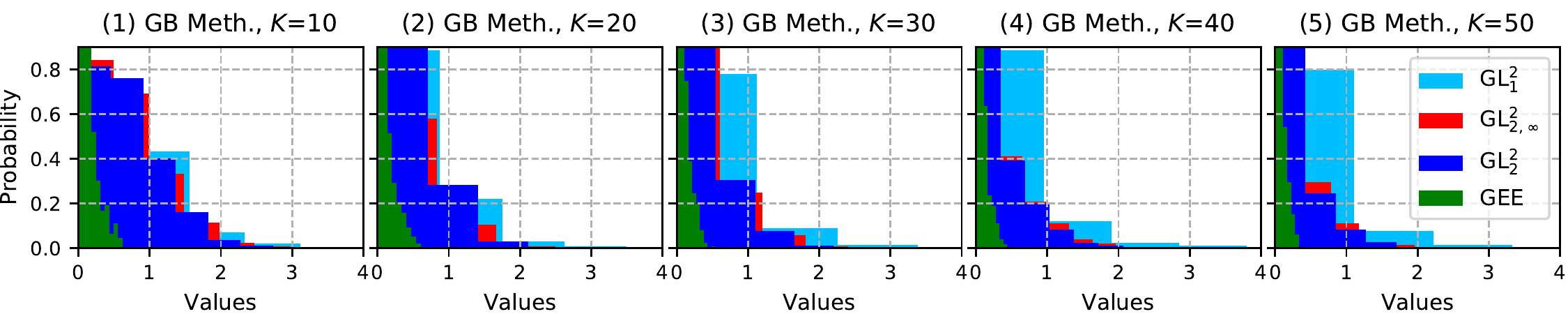} \label{fig:nmf_regularizers_distributions_gene_body}}\vspace{-0.1em}
\caption{Convergence of the models on the 
Gene body methylation dataset (upper) and the distribution of factored $\bW$ (lower), measuring the training data fit (mean squared error). When we increase latent dimension $K$, all the models continue to increase the performance.}
\label{fig:convergences_ctrp_nmf_regularizer}
\end{figure*}

\section{Experiments}\label{section:nmf_regulari_experiments}

\begin{figure}[H]
	\centering  
	\vspace{-0.35cm} 
	\subfigtopskip=2pt 
	\subfigbottomskip=2pt 
	\subfigcapskip=-5pt 
	\subfigure{\includegraphics[width=0.231\textwidth]{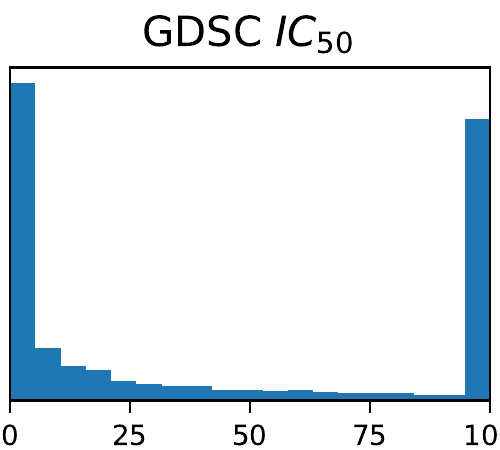} \label{fig:plot_gdsc}}
	\subfigure{\includegraphics[width=0.231\textwidth]{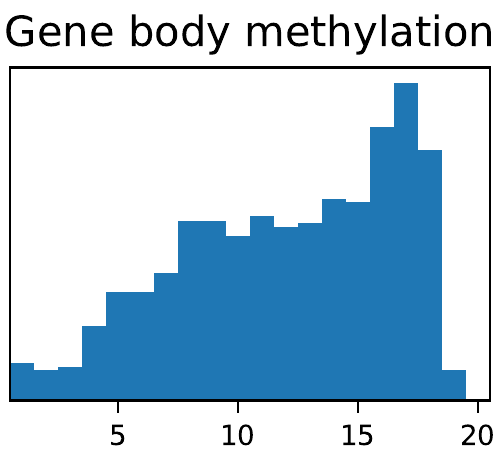} \label{fig:plot_methylation_gm}}
	\caption{Data distribution of GDSC $IC_{50}$ and Gene body methylation datasets.}
	\label{fig:datasets_nmf_regularizer}
\end{figure}

We conduct experiments with various analysis tasks to demonstrate the main advantages of the proposed GL$_2^2$ and GL$_{2,\infty}^2$ methods. We use two datasets from bioinformatics: The first one is the Genomics of Drug Sensitivity in Cancer dataset\footnote{\url{https://www.cancerrxgene.org/}} (GDSC $IC_{50}$) \citep{yang2012genomics}, which contains a wide range of drugs and their treatment outcomes on different cancer and tissue types (cell lines).
Following \citet{brouwer2017prior}, we preprocess the GDSC $IC_{50}$ dataset by capping high values
to 100, undoing the natural log transform, and casting them as integers. 
The second one is the Gene body methylation dataset \citep{koboldt2012comprehensive}, which gives the amount of methylation measured in the body region of 160 breast cancer driver genes.
We multiply the values in Gene body methylation dataset by 20 and cast them as integers as well.
A summary of the two datasets can be seen in Table~\ref{table:datadescription_nmf_regularizer} and their distributions are shown in Figure~\ref{fig:datasets_nmf_regularizer}.
The GDSC $IC_{50}$ dataset has a larger range whose values are unbalanced (either small as 0 or larger as 100);
while the Gene body methylation dataset has a smaller range whose values seem balanced.
We can see that the GDSC $IC_{50}$ is relatively a large dataset whose matrix rank is $139$ and the Gene body methylation data tends to be small whose matrix rank is 160. 

\begin{table}[h]
\setlength{\tabcolsep}{6.9pt}
	\begin{tabular}{l|lll}
		\hline
		Dataset        & Rows & Columns & Fraction obs. \\ \hline
		GDSC $IC_{50}$ & 707  & 139     & 0.806         \\
		Gene body meth.  & 160  & 254     & 1.000         \\
		\hline
	\end{tabular}
	\caption{Dataset description. Gene body methylation is relatively a small dataset and the GDSC $IC_{50}$ tends to be large. The description provides the number of rows, columns, and the fraction of entries that are observed.}
	\label{table:datadescription_nmf_regularizer}
\end{table}

The same parameter initialization is adopted in each scenario. 
We compare the results in terms of convergence speed and generalization. In a wide range of scenarios across various models, GL$_2^2$ and GL$_{2,\infty}^2$ improve convergence rates, and lead to out-of-sample performance that is as good or better than existing Bayesian NMF models.

\subsection{Hyperparameters}
We follow the default hyperparameter setups in \citet{brouwer2017prior}. We use $\{\lambda_{mk}^W\}=\{\lambda_{kn}^Z\}=0.1$ (GEE); 
$\{\lambda_{k}^W\}=\{\lambda_{k}^Z\}=0.1$ (GL$_1^2$, GL$_2^2$, GL$_{2,\infty}^2$);
uninformative $\alpha_\sigma=\beta_\sigma=1$ (inverse-Gamma prior in GEE, GL$_1^2$, GL$_2^2$, GL$_{2,\infty}^2$).
These are very weak prior choices and the models are insensitive to them \citep{brouwer2017prior}.
As long as the hyperparameters are set, the observed or unobserved variables are initialized from random draws as this initialization procedure provides a better initial guess of the right patterns in the matrices.
In all experiments, we run the Gibbs sampler 500 iterations with a burn-in of 300 iterations as the convergence analysis shows the algorithm can converge in fewer than 200 iterations.

\subsection{Convergence Analysis}
\paragraph{GDSC $IC_{50}$ with relatively large entries} Firstly we compare the convergence in terms of iterations on the GDSC $IC_{50}$  and Gene body methylation datasets. We run each model with $K=\{10, 20, 30, 40, 50\}$, and the loss is measured by mean squared error (MSE).
Figure~\ref{fig:nmf_regularizers_convergences_gdsc}  shows the average convergence results of ten repeats and 
Figure~\ref{fig:nmf_regularizers_distributions_gdsc} shows the distribution of entries of the factored $\bW$ for the last 20 iterations
 on the GDSC $IC_{50}$ dataset. 
The result is consistent with our analysis (Section~\ref{section:gibbs_nmf_regu}, the connection between different models). Since the values of the data matrix for GDSC $IC_{50}$ dataset is large, the posterior ``parent" mean $\widetilde{\mu_{mk}}$ in GL$_1^2$ model is approaching zero or even negative, thus it has a larger regularization than GEE model. This makes the GL$_1^2$ model converge to a worse performance. GL$_2^2$ and GL$_{2,\infty}^2$ models, on the contrary, impose looser regularization than the GL$_1^2$ model, and the convergence performances are close to that of the GEE model.

\paragraph{Gene body methylation with relatively small entries}
Figure~\ref{fig:nmf_regularizers_convergences_gene_body} further shows the average convergence results of ten repeats, and Figure~\ref{fig:nmf_regularizers_distributions_gene_body} shows the distribution of the entries of the factored $\bW$ for the last 20 iterations on the Gene body methylation dataset. 
The situation is different for the GL$_1^2$ model since the range of the entries of the Gene body methylation dataset is smaller than that of the GDSC $IC_{50}$ dataset (see Figure~\ref{fig:datasets_nmf_regularizer}).
This makes the $-\lambda_k^W\cdot\textcolor{black}{\sum_{j\neq k}^{K}w_{mj}}$ term of posterior ``parent" mean $\widetilde{\mu_{mk}}$ in GL$_1^2$ model approach zero (see Table~\ref{table:nmf_regularizer_posterior}), and the model then favors a looser regularization than the GEE model. 

The situation can be further presented by the distribution of the factored component $\bW$ on the GDSC $IC_{50}$ (Figure~\ref{fig:nmf_regularizers_distributions_gdsc}) and on the Gene body methylation (Figure~\ref{fig:nmf_regularizers_distributions_gene_body}). GEE model has larger values of $\bW$ on the former dataset and smaller values on the latter; while GL$_1^2$ has smaller values of $\bW$ on the former dataset and larger values on the latter.
In other words, the regularization of the GEE and GL$_1^2$ is inconsistent on the two different data matrices.
In comparison, the proposed GL$_2^2$ and GL$_{2,\infty}^2$   are consistent on different datasets, making them more robust algorithms to compute NMF.


\begin{table}[h]
\setlength{\tabcolsep}{6.8pt}
\begin{tabular}{l|llll}
\hline
$K$ & GEE & GL$_1^2$   & GL$_2^2$ & GL$_{2,\infty}^2$  \\ \hline
10   &  8.1 (1.9)   &  1.3 (10.3)    &    2.4 (3.8) &   2.4 (4.5)  \\
20   &  8.6 (1.5)   &  0.8 (14.7)    &    2.3 (4.1) &   2.2 (4.4)  \\
30   &  8.7 (1.4)   &  0.7 (17.3)    &    2.2 (4.3) &   2.2 (4.4)  \\
40   &  8.3 (1.5)   &  0.6 (19.4)    &    2.2 (4.4) &   2.2 (4.4)  \\
50   &  8.0 (1.6)   &  0.5 (21.2)    &    2.2 (4.1) &   2.2 (4.2)  \\
\hline
\hline
10   &  0.1 (80.4)   &  0.7 (11.4)    &    0.7 (11.5) &   0.7 (12.7)  \\
20   &  0.1 (87.8)   &  0.6 (16.2)    &    0.5 (21.3) &   0.5 (21.0)  \\
30   &  0.0 (90.2)   &  0.6 (18.2)    &    0.3 (37.1) &   0.3 (36.4)  \\
40   &  0.0 (92.2)   &  0.6 (20.8)    &    0.3 (48.9) &   0.3 (49.1)  \\
50   &  0.0 (93.0)   &  0.5 (22.8)    &    0.2 (58.4) &   0.2 (58.4)  \\
\hline
\end{tabular}
\caption{Mean values of the factored component $\bW$ in the last 20 iterations, where the value in the (parentheses) is the sparsity evaluated by taking the percentage of values smaller than 0.1, for GDSC $IC_{50}$ (upper table) and  Gene body methylation (lower table).
The inconsistency of GEE and GL$_1^2$ for different matrices can be observed.
}
\label{table:nmfreg_distribtuionsvalues}
\end{table}

\begin{table}[h]
\setlength{\tabcolsep}{5.3pt}
\begin{tabular}{l|l|llll}
\hline

Unobs.& $K$ & GEE & GL$_1^2$   & GL$_2^2$ & GL$_{2,\infty}^2$   \\ \hline
\parbox[t]{5.0mm}{\multirow{4}{*}{\rotatebox[origin=c]{0}{60\%}  }} 
&20   &  787.60   &  880.36    &   \textbf{ 769.24  } &   \textbf{ 768.27 } \\
&30   &  810.39   &  888.47    &   \textbf{ 774.53  } &   \textbf{ 773.27 } \\
&40   &  802.39   &  892.01    &   \textbf{ 783.26  } &   \textbf{ 784.30 } \\
&50   &  \textbf{795.72}   &  895.05    &   \textbf{ 806.14  } &   \textbf{ 807.44 } \\
\hline
\hline
\parbox[t]{5.0mm}{\multirow{4}{*}{\rotatebox[origin=c]{0}{70\%}  }} 
&20   &  841.74   &  895.77    &   \textbf{ 798.44  } &   \textbf{ 796.15 } \\
&30   &  830.45   &  902.48    &   \textbf{ 807.37  } &   \textbf{ 806.61 } \\
&40   &  \textbf{842.70}   &  907.65    &   \textbf{ 832.67  } &   \textbf{ 835.89 } \\
&50   & \textbf{846.83}   &  {1018.97}	$\uparrow$    &   \textbf{ 864.58  } &   \textbf{ 869.15 } \\
\hline
\hline
\parbox[t]{5.0mm}{\multirow{4}{*}{\rotatebox[origin=c]{0}{80\%}  }} 
&20   &  904.39   &  926.72    &   \textbf{ 842.24  } &   \textbf{ 841.84 } \\
&30   &  \textbf{887.63}   &  938.92    &   \textbf{ 879.30  } &   \textbf{ 883.57 } \\
&40   &  \textbf{942.44}   &  2634.69    &   \textbf{ 935.09  } &   \textbf{ 939.77 } \\
&50   &  \textbf{952.45}   &  {2730.30} $\uparrow$     &   \textbf{ 974.01  } &   \textbf{ 973.75 } \\
\hline
\end{tabular}
\caption{Mean squared error measure when the percentage of unobserved data is 60\% (upper table), 70\% (middle table), or 80\% (lower table) for the GDSC $IC_{50}$ dataset. The performance of the proposed GL$_2^2$ and GL$_{2,\infty}^2$ models is only slightly worse when we increase the fraction of unobserved from 60\% to 80\%; while the performance of GL$_1^2$ becomes extremely poor.
Similar observations occur in the Gene body methylation experiment. The symbol $\uparrow$ means the performance becomes extremely worse.}
\label{table:nmfregu_special_sparsity_case}
\end{table}

\begin{figure*}[h]
\centering  
\subfigtopskip=2pt 
\subfigbottomskip=2pt 
\subfigcapskip=-2pt 
\subfigure[Predictive results on the \textbf{GDSC $\boldsymbol{IC_{50}}$} dataset with increasing fraction of unobserved data and increasing latent dimension $K$.]{\includegraphics[width=1\textwidth]{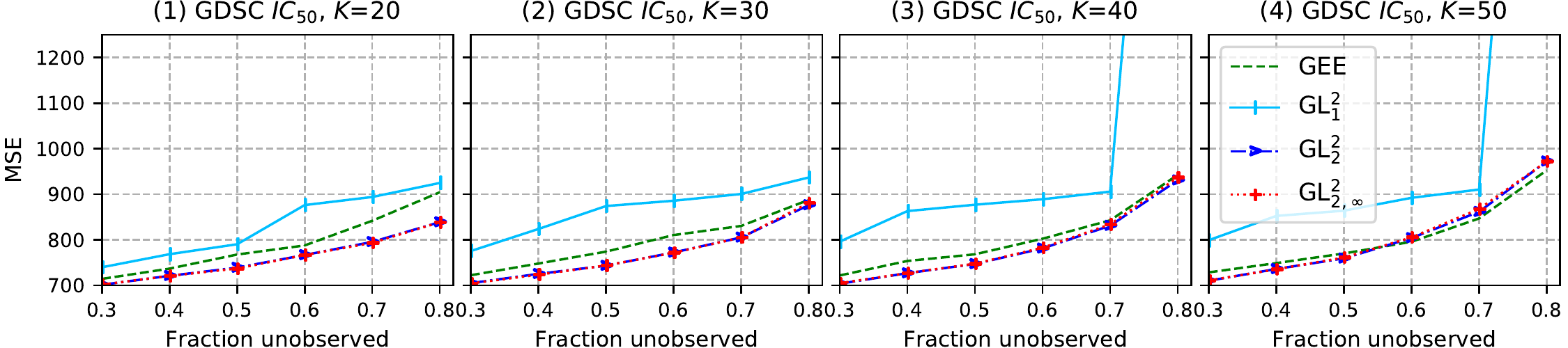} 
\label{fig:nmf_regularizer_sparsity_movielens_gdsc}}
\subfigure[Predictive results on \textbf{Gene body methylation} dataset  with increasing fraction of unobserved data and increasing latent dimension $K$.]{\includegraphics[width=1\textwidth]{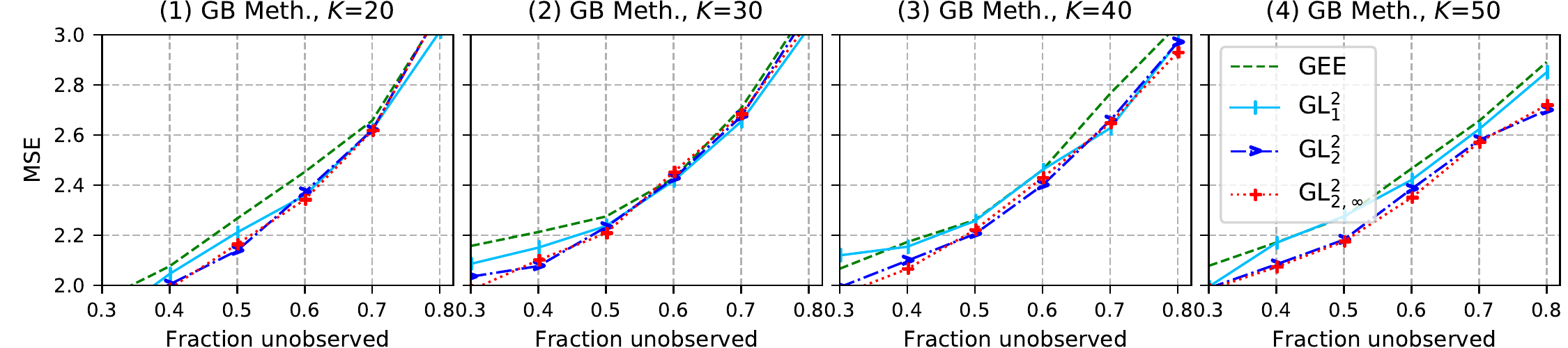} 
\label{fig:nmf_regularizer_sparsity_movielens_genebody_meth}}
\caption{Predictive results on the \textbf{GDSC $\boldsymbol{IC_{50}}$} (upper) and \textbf{Gene body methylation} (lower) datasets.
We measure the predictive performance (mean squared error) on a held-out dataset for different fractions of unobserved data. }
	\label{fig:sparsity_nmf_regulariza}
\end{figure*}

Table~\ref{table:nmfreg_distribtuionsvalues} shows the mean values of the factored component $\bW$ in the last 20 iterations for GDSC $IC_{50}$ (upper table) and  Gene body methylation (lower table) where the value in the parentheses is the sparsity evaluated by taking the percentage of values smaller than 0.1.
The inconsistency of GEE and GL$_1^2$ for different matrices can be observed (either large sparsity or small sparsity), while the results for the proposed GL$_2^2$ and GL$_{2,\infty}^2$ models are more consistent.

\subsection{Predictive Analysis}

The training performances of the GEE, GL$_2^2$, and GL$_{2,\infty}^2$ models steadily improve as the model complexity grows. Inspired by this result, 
we measure the predictive performance when the sparsity of the data increases to see whether the models overfit or not. For different fractions of unobserved data, we randomly split the data based on that fraction, train the model on the observed data, and measure the performance on the held-out test data. Again, we increase $K$ from $K=20$ to $K=30, 40, 50$ for all models. The average MSE of ten repeats is given in Figure~\ref{fig:sparsity_nmf_regulariza}. 
We still observe the inconsistency issue in the GL$_1^2$ model, the predictive performance of it is as good as that of the proposed GL$_2^2$ and GL$_{2,\infty}^2$ models on the Gene body methylation dataset; while the predictive results of the GL$_1^2$ model are extremely poor on the GDSC $IC_{50}$ dataset. 

For the GDSC $IC_{50}$ dataset, the proposed GL$_2^2$ and GL$_{2,\infty}^2$ models perform best when the latent dimensions are $K=20, 30, 40$; when $K=50$ and the fraction of unobserved data increases, the GEE model is slightly better. As aforementioned, the GL$_1^2$ performs the worst on this dataset; and when the fraction of unobserved data increases or $K$ increases, the predictive results of GL$_1^2$ deteriorate quickly.

For the Gene body methylation dataset, the predictive performance of GL$_1^2$, GL$_2^2$ and GL$_{2,\infty}^2$ models are close (GL$_1^2$ has a slightly larger error). The GEE model performs the worst on this dataset.

The comparison on the two sets shows the proposed GL$_2^2$ and GL$_{2,\infty}^2$ models have both better in-sample and out-of-sample performance, making them a more robust choice in predicting missing entries.

Table~\ref{table:nmfregu_special_sparsity_case} shows MSE predictions of different models when the fractions of unobserved data is $60\%$, $70\%$, and $80\%$ respectively. We observe that the performances of the proposed GL$_2^2$ and GL$_{2,\infty}^2$ models are only slightly worse when we increase the fraction of unobserved from 60\% to 80\%. This indicates the proposed GL$_2^2$ and GL$_{2,\infty}^2$ models are more robust with less overfitting. While for the GL$_1^2$ model, the performance becomes extremely poor in this scenario.



\subsection{Noise Sensitivity}
\begin{figure}[h]
	\centering  
	\subfigtopskip=2pt 
	\subfigbottomskip=9pt 
	\subfigcapskip=-5pt 
	\includegraphics[width=0.33\textwidth]{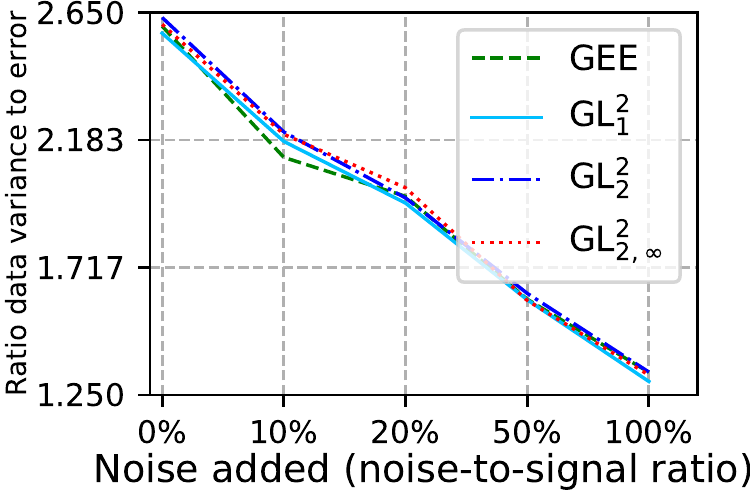}
	\caption{Ratio of the variance of data to the MSE of the predictions, the higher the better.}
	\label{fig:noise_graph_gdsc}
\end{figure}

Finally, we measure the noise sensitivity of different models with predictive performance when the datasets are noisy. To see this, we add different levels of Gaussian noise to the data. We add levels of 
$\{0\%, 10\%,$ $20\%,$ $50\%, 100\%\}$ 
noise-to-signal ratio noise (which is the ratio of the variance of the added Gaussian noise to the variance of the data). The results for the GDSC $IC_{50}$ with $K=10$ are shown in Figure~\ref{fig:noise_graph_gdsc}. The results are the average performance over 10 repeats. We observe that the proposed GL$_2^2$ and GL$_{2,\infty}^2$ models perform slightly better than other NMF models.  
The proposed GL$_2^2$ and GL$_{2,\infty}^2$ models perform notably better when the noise-to-signal ratio is smaller than 10\% and slightly better when the ratio is larger than 20\%.
Similar results can be found on the Gene body methylation dataset and other $K$ values and we shall not repeat the details.

\section{Conclusion}
This article proposes a general framework of Bayesian NMF with implicit norm regularization.
This is a simple and computationally efficient algorithm that requires no extra computation and is easy to implement for nonnegative matrix factorization. Overall, we show that the proposed GL$_2^2$ and GL$_{2,\infty}^2$ models are versatile algorithms that have better convergence results and out-of-sample performance on datasets with both small and large range of values. 
GL$_2^2$ and GL$_{2,\infty}^2$ are able to avoid problems of overfitting, which are common in the standard non-probabilistic NMF model \citep{schmidt2009probabilistic} and other Bayesian NMF models \citep{brouwer2017prior}.

\small
\bibliography{bib}
\bibliographystyle{sty}
\balance

\onecolumn
\appendix

\normalsize
\section{Gibbs Sampling Algorithms of NMF Models}\label{appendix:gibbs_NMF_regularizer}
We give the posteriors for several Bayesian NMF models that we compare in the paper, namely, the GEE, GL$_1^2$, GL$_2^2$, and GL$_\infty$ models. The derivation for GL$_{2,\infty}^2$ model is just the same as the GL$_2^2$ and GL$_\infty$ models. For clarity, the parameters for posterior densities are denoted using symbols with a widetilde, e.g., $\widetilde{\mu}$ and $\widetilde{\sigma^2}$ for the posterior mean and variance respectively.

\subsection{Gaussian Likelihood with Exponential Priors (GEE) Model}

\begin{figure}[h]
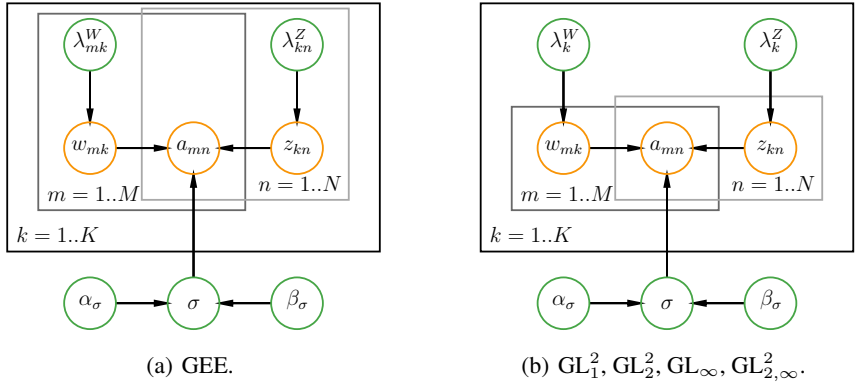

	\centering  
	\vspace{-0.35cm} 
	\subfigtopskip=2pt 
	\subfigbottomskip=2pt 
	\subfigcapskip=-5pt 
	\subfigure[GEE.]{\label{fig:bmf_gee_append}
		\includegraphics[width=0.321\linewidth]{./imgs/bmf_gee.pdf}}
	\hspace{1.5em}
	\subfigure[GL$_1^2$, GL$_2^2$, GL$_\infty$, GL$_{2,\infty}^2$.]{\label{fig:bmf_gtt_append}
		\includegraphics[width=0.321\linewidth]{./imgs/bmf_gl12.pdf}}
	\caption{\textbf{Same as Figure~\ref{fig:bmf_gl12_gee_nmf_regu} with higher resolution.} Graphical model representation of GEE, GL$_1^2$, GL$_2^2$, GL$_\infty$, and GL$_{2,\infty}^2$ models. Orange circles represent observed and latent variables, green circles denote prior variables, and plates represent repeated variables.}
	\label{fig:bmf_nmf_regularizer_graphical_pms}
\end{figure}

The Gaussian Exponential-Exponential (GEE) model is perhaps the simplest one for Bayesian NMF \citep{schmidt2009bayesian} where each entry $a_{mn}$ of matrix $\bA$ is again modeled using a Gaussian likelihood with variance $\sigma^2$ and mean given by the latent decomposition $\bw_m^\top\bz_n$ (Eq.~\eqref{equation:grrn_data_entry_likelihood}, this will be default for  GL$_1^2$, GL$_2^2$, GL$_\infty$, and GL$_{2,\infty}^2$ models as well). The graphical representation is shown in Figure~\ref{fig:bmf_gee_append}:

The model places independent exponential priors over the entries of $\bW, \bZ$, 
$$
\begin{aligned}
w_{mk}&\sim \exponential(w_{mk}| \lambda_{mk}^W), \qquad 
&z_{kn}&\sim \exponential(z_{kn}| \lambda_{kn}^W);\\
p(\bW|\{\lambda_{mk}^W \}) &=\prod_{m,k=1}^{M,K} \exponential(w_{mk}| \lambda_{mk}^W), \qquad\qquad
&p(\bZ|\{\lambda_{kn}^Z \}) &=\prod_{k,n=1}^{K,N} \exponential(z_{kn}| \lambda_{kn}^Z). 
\end{aligned}
$$
Denote $\blambda^W$ as the $M\times K$ matrix containing all $\{\lambda_{mk}^W\}$ entries, $\blambda^Z$ as the $K\times N$ matrix including all $\{\lambda_{kn}^Z\}$ values, and $\bW_{-{mk}}$ as all elements of $\bW$ except $w_{mk}$. 
The product of a Gaussian and an exponential distribution leads to a truncated-normal posterior,
\begin{equation}\label{equation:gee_poster_wmk1_append}
\begin{aligned}
&\gap p(w_{mk}| \sigma^2, \bW_{-mk}, \bZ, \blambda^W,\cancel{\blambda^Z}, \bA)=p(w_{mk} | \sigma^2,   \bW_{-mk}, \bZ, \lambda_{mk}^W, \bA) \\
&\propto p(\bA| \bW, \bZ, \sigma^2) \times p(w_{mk}| \lambda_{mk}^W)
=\prod_{i,j=1}^{M,N} \normal \left(a_{ij}| \bw_i^\top\bz_j, \sigma^2 \right)\times \exponential(w_{mk}| \lambda_{mk}^W) \\
&\propto \exp\left\{   -\frac{1}{2\sigma^2}  \sum_{i,j=1}^{M,N}(a_{ij} - \bw_i^\top\bz_j  )^2\right\}  \times \cancel{\lambda_{mk}^W }\exp(-\lambda_{mk}^W \cdot w_{mk})u(w_{mk})\\
&\propto \exp\left\{   -\frac{1}{2\sigma^2}  \sum_{j=1}^{N}(a_{mj} - \bw_m^\top\bz_j  )^2\right\}  \cdot  \exp(-\lambda_{mk}^W\cdot w_{mk})u(w_{mk})\\
&\propto \exp\left\{   -\frac{1}{2\sigma^2}  \sum_{j=1}^{N}
\left( w_{mk}^2z_{kj}^2 + 2w_{mk} z_{kj}\bigg(\sum_{i\neq k}^{K}w_{mi}z_{ij} - a_{mj}\bigg)  \right)
\right\}  \cdot \exp(-\lambda_{mk}^W\cdot w_{mk})u(w_{mk})\\
&\propto \exp\left\{   
-\underbrace{\left(\frac{\sum_{j=1}^{N} z_{kj}^2 }{2\sigma^2}  \right) }_{\textcolor{blue}{1/(2\widetilde{\sigma_{mk}^{2}})}}
w_{mk}^2 
+
w_{mk}\underbrace{\left( -\lambda_{mk}^W+ \frac{1}{\sigma^2} \sum_{j=1}^{N} z_{kj}\bigg( a_{mj} - \sum_{i\neq k}^{K}w_{mi}z_{ij}\bigg)  \right)}_{\textcolor{blue}{\widetilde{\sigma_{mk}^{2}}^{-1} \widetilde{\mu_{mk}}}}
\right\}  \cdot u(w_{mk})\\
&\propto   \normal(w_{mk} | \widetilde{\mu_{mk}}, \widetilde{\sigma_{mk}^{2}})\cdot u(w_{mk}) 
= \truncatednormal(w_{mk} | \widetilde{\mu_{mk}}, \widetilde{\sigma_{mk}^{2}}),
\end{aligned}
\end{equation}
where 
\begin{equation}\label{equation:gee_posterior_variance_append}
	\widetilde{\sigma_{mk}^{2}}= \frac{\sigma^2}{\sum_{j=1}^{N} z_{kj}^2}
\end{equation}
is the posterior variance of the normal distribution with mean $\widetilde{\mu_{mk}}$, 
\begin{equation}\label{equation:gee_posterior_mean_append}
	\widetilde{\mu_{mk}} = \left( -\lambda_{mk}^W+ \frac{1}{\sigma^2} \sum_{j=1}^{N} z_{kj}\bigg( a_{mj} - \sum_{i\neq k}^{K}w_{mi}z_{ij}\bigg)  \right)\cdot \widetilde{\sigma_{mk}^{2}}
\end{equation}
is the posterior mean of the normal distribution and $\truncatednormal(x | \mu, \sigma^2)$ is the \textit{truncated normal density} with ``parent" mean $\mu$ and ``parent" variance $\sigma^2$.

Or after rearrangement, the posterior density of $w_{mk}$ can be equivalently described by 
$$
p(w_{mk} | \sigma^2, \bW_{-mk}, \bZ, \lambda, \bA) 
=  \rectifieddist(w_{mk} |   \widehat{\mu_{mk}}, \widehat{\sigma_{mk}^{2}}, \lambda_{mk}^W ),
$$
where $\widehat{\sigma^2_{mk}}=\widetilde{\sigma_{mk}^{2}}= \frac{\sigma^2}{\sum_{j=1}^{N} z_{kj}^2}$ is the posterior ``parent" variance of the normal distribution with ``parent" mean $\widehat{\mu_{mk}}$, 
$$
\widehat{\mu_{mk}} =  \frac{1}{\sum_{j=1}^{N} z_{kj}^2} \cdot \sum_{j=1}^{N} z_{kj}\bigg( a_{mj} - \sum_{i\neq k}^{K}w_{mi}z_{ij}\bigg) .
$$
Due to symmetry, a similar expression for $z_{kn}$ can be easily derived.

The conditional density of $\sigma^2$ depends on its parents ($\alpha_\sigma$, $\beta_\sigma$), children ($\bA$), and co-parents ($\bW$, $\bZ$) in the graph. And it is an inverse-gamma distribution (by conjugacy), 
\begin{equation}\label{equation:gee_posterior_sigma2_append}
\begin{aligned}
&p(\sigma^2 | {\bW}, {\bZ}, \cancel{\blambda^W}, \cancel{\blambda^Z}, \bA)=
p(\sigma^2 | \bW,\bZ, \bA) = \inversegammadist (\sigma^2| \widetilde{\alpha_{\sigma}}, \widetilde{\beta_{\sigma}}), \gap\gap\qquad \\
& \widetilde{\alpha_{\sigma}} = \frac{MN}{2} +{\alpha_\sigma}, \gap \gap\gap
\widetilde{\beta_{\sigma}}  =  \frac{1}{2} \sum_{m,n=1}^{M,N} (\bA-\bW\bZ)_{mn}^2 + {\beta_\sigma}.
\end{aligned}
\end{equation}

\subsection{Gaussian Likelihood with $L_1^2$ Prior (GL$_1^2$) Model}
The Gaussian $L_1^2$ Norm Model (GL$_1^2$) model is proposed by \citet{brouwer2017prior} based on the $L_1^2$ norm for both $\bW, \bZ$.

\paragraph{Prior} We assume $\bW$ and $\bZ$ are independently distributed with parameter $\lambda_{k}^W$ and $\lambda_{k}^Z$ proportional to a exponential function: 
\begin{equation}\label{equation:gl12_prior_density_append}
\begin{aligned}
&p(\bW|\lambda_{k}^W) &\propto &
\left\{
\begin{aligned}
&\exp \left[  -\frac{\lambda_{k}^W}{2} \sum_{m=1}^{M} \left(\sum_{k=1}^{K} w_{mk}\right)^2 \right]
, &\gap &\text{if $w_{mk}\geq 0$ for all $m,k$ };
\\
&0, &\gap &\text{if otherwise};
\end{aligned}
\right.\\
\gap 
&p(\bZ|\lambda_{k}^Z) &\propto &
\left\{
\begin{aligned}
&\exp \left[  -\frac{\lambda_{k}^Z}{2} \sum_{n=1}^{N} \left(\sum_{k=1}^{K} z_{kn}\right)^2 \right]
, &\gap &\text{if $z_{kn}\geq 0$ for all $n,k$ };
\\
&0, &\gap &\text{if otherwise}.
\end{aligned}
\right.\\
\end{aligned}
\end{equation}
Again, the prior for the noise variance $\sigma^2$ is an inverse-gamma density with shape ${\alpha_\sigma}$ and scale ${\beta_\sigma}$,
$$
p(\sigma^2)= \inversegammadist(\sigma^2| \alpha_\sigma, \beta_\sigma) = \frac{{\beta_\sigma}^{\alpha_\sigma}}{\Gamma({\alpha_\sigma})} (\sigma^2)^{-\alpha_\sigma-1} \exp\left( -\frac{{\beta_\sigma}}{\sigma^2} \right).
$$
\paragraph{Posterior}
The conditional density of $\sigma^2$ is the same as that in the GEE model (Eq.~\eqref{equation:gee_posterior_sigma2_append}). 
By Bayes' rule, the posterior is proportional to the product of likelihood and prior, it can be maximized to yield an estimate of $\bW$ and $\bZ$:
\begin{equation}\label{equation:gl12_poster_wmk1_append}
\begin{aligned}
&\gap p(w_{mk}| \sigma^2, \bW_{-mk}, \bZ, \lambda_{k}^W,\cancel{\lambda_{k}^Z}, \bA)=p(w_{mk} | \sigma^2,   \bW_{-mk}, \bZ, \lambda_{k}^W, \bA) \\
&\propto p(\bA| \bW, \bZ, \sigma^2) \times p(\bW| \lambda_{k}^W)
=\prod_{i,j=1}^{M,N} \normal \left(a_{ij}| \bw_i^\top\bz_j, \sigma^2 \right)\times 
p(\bW| \lambda_{k}^W)\\
&\propto \exp\left\{   -\frac{1}{2\sigma^2}  \sum_{i,j=1}^{M,N}(a_{ij} - \bw_i^\top\bz_j  )^2\right\}  \times 
\exp \left\{ -\frac{\lambda_{k}^W}{2} \sum_{i=1}^{M} \left(\sum_{j=1}^{K} w_{ij}\right)^2 \right\} \cdot u(w_{mk}) \\
&\propto \exp\left\{   -\frac{1}{2\sigma^2}  \sum_{j=1}^{N}(a_{mj} - \bw_m^\top\bz_j  )^2\right\}  \times   
\exp\left\{ -\frac{\lambda_{k}^W}{2} \left( w_{mk} + \sum_{j\neq k }^{K}w_{mj}\right)^2 \right\}\cdot u(w_{mk}) \\
&\propto \exp\left\{   -\frac{1}{2\sigma^2}  \sum_{j=1}^{N}
\left( w_{mk}^2z_{kj}^2 + 2w_{mk} z_{kj}\bigg(\sum_{i\neq k}^{K}w_{mi}z_{ij} - a_{mj}\bigg)  \right)
\right\}  \exp\left\{ \frac{-\lambda^W}{2}  w_{mk}^2 - \lambda_{k}^W  w_{mk}  \sum_{j\neq k }^{K}w_{mj}
\right\}  u(w_{mk})\\
&\propto \exp\Bigg\{   
-
\underbrace{\left(\frac{\sum_{j=1}^{N} z_{kj}^2 + \textcolor{red}{\sigma^2\lambda_{k}^W} }{2\sigma^2}  \right)  }_{\textcolor{blue}{ 1/(2\widetilde{\sigma^2_{mk} }) }} 
w_{mk}^2+ w_{mk}\underbrace{\left( -\lambda_{k}^W \cdot \textcolor{red}{\sum_{j\neq k}^{K}w_{mj}}+ 
\frac{1}{\sigma^2} \sum_{j=1}^{N} z_{kj}\bigg( a_{mj} - \sum_{i\neq k}^{K}w_{mi}z_{ij}\bigg)  \right)}_{\textcolor{blue}{\widetilde{\sigma_{mk}^{2}}^{-1} \widetilde{\mu_{mk}}}}
\Bigg\}  \cdot u(w_{mk})\\
&\propto   \normal(w_{mk} | \widetilde{\mu_{mk}}, \widetilde{\sigma_{mk}^{2}})\cdot u(w_{mk}) 
= \truncatednormal(w_{mk} | \widetilde{\mu_{mk}}, \widetilde{\sigma_{mk}^{2}}),
\end{aligned}
\end{equation}
where 
$$
\begin{aligned}
\widetilde{\sigma_{mk}^{2}}= \frac{\sigma^2}{\sum_{j=1}^{N} z_{kj}^2 +\textcolor{red}{\sigma^2\lambda_{k}^W}} , \gap \gap 
\widetilde{\mu_{mk}} = \left( -\lambda_{k}^W\cdot\textcolor{red}{\sum_{j\neq k}^{K}w_{mj}}
+
\frac{1}{\sigma^2} \sum_{j=1}^{N} z_{kj}\bigg( a_{mj} - \sum_{i\neq k}^{K}w_{mi}z_{ij}\bigg)  \right)\cdot \widetilde{\sigma_{mk}^{2}}
\end{aligned}
$$ 
are the posterior ``parent"  variance of the normal distribution, and the posterior ``parent" mean of the normal distribution respectively. Note the posterior density of $w_{mk}$ in Eq.~\eqref{equation:gl12_poster_wmk1_append} is very similar to that of the GEE model in Eq.~\eqref{equation:gee_poster_wmk1_append} where we highlight the difference in \textcolor{red}{red} text.

\paragraph{Connection to GEE}
The second term $\sigma^2\lambda_k^W$ exists in the GL$_1^2$ denominator of the variance $\widetilde{\sigma_{mk}^{2}}$. When all else are held equal, the GL$_1^2$ has smaller variance than GEE, so the distribution of GL$_1^2$ is more clustered in a smaller range. This is actually a stronger constraint/regularizer than GEE model. 


\subsection{Gaussian Likelihood with $L_2^2$ Prior (GL$_2^2$) Model}
In the main paper, we further propose the GL$_2^2$ model based on $L_2$ norm.

\paragraph{Prior} We assume $\bW$ and $\bZ$ are independently distributed with parameter $\lambda_{k}^W$  and $\lambda_{k}^Z$ proportional to a exponential function: 
\begin{equation}\label{equation:gp22_prior_density_append}
	\begin{aligned}
		&p(\bW|\lambda_k^W) &\propto &
		\left\{
		\begin{aligned}
			&\exp \left[  -\frac{\lambda_k^W}{2} \sum_{m=1}^{M} \left(\sum_{k=1}^{K} w_{mk}^2\right) \right]
			, &\gap &\text{if $w_{mk}\geq 0$ for all $m,k$ };
			\\
			&0, &\gap &\text{if otherwise};
		\end{aligned}
		\right.\\
		\gap 
		&p(\bZ|\lambda_k^Z) &\propto &
		\left\{
		\begin{aligned}
			&\exp \left[  -\frac{\lambda_k^Z}{2} \sum_{n=1}^{N} \left(\sum_{k=1}^{K} z_{kn}^2\right) \right]
			, &\gap &\text{if $z_{kn}\geq 0$ for all $n,k$ };
			\\
			&0, &\gap &\text{if otherwise}.
		\end{aligned}
		\right.\\
	\end{aligned}
\end{equation}
Again, the prior for the noise variance $\sigma^2$ is an inverse-gamma density with shape ${\alpha_\sigma}$ and scale ${\beta_\sigma}$,
$$
p(\sigma^2)= \inversegammadist(\sigma^2| \alpha_\sigma, \beta_\sigma) = \frac{{\beta_\sigma}^{\alpha_\sigma}}{\Gamma({\alpha_\sigma})} (\sigma^2)^{-\alpha_\sigma-1} \exp\left( -\frac{{\beta_\sigma}}{\sigma^2} \right).
$$
\paragraph{Posterior}
The conditional density of $\sigma^2$ is the same as that in the GEE model (Eq.~\eqref{equation:gee_posterior_sigma2_append}). 
By Bayes' rule, the posterior is proportional to the product of likelihood and prior, it can be maximized to yield an estimate of $\bW$ and $\bZ$:
\begin{equation}\label{equation:gp22_poster_wmk1_append}
\begin{aligned}
&\gap p(w_{mk}| \sigma^2, \bW_{-mk}, \bZ, \lambda_k^W,\cancel{\lambda_k^Z}, \bA)=p(w_{mk} | \sigma^2,   \bW_{-mk}, \bZ, \lambda_k^W, \bA) \\
&\propto p(\bA| \bW, \bZ, \sigma^2) \times p(\bW| \lambda_k^W)
=\prod_{i,j=1}^{M,N} \normal \left(a_{ij}| \bw_i^\top\bz_j, \sigma^2 \right)\times 
p(\bW| \lambda_k^W)
\cdot u(w_{mk}) \\
&\propto \exp\left\{   -\frac{1}{2\sigma^2}  \sum_{i,j=1}^{M,N}(a_{ij} - \bw_i^\top\bz_j  )^2\right\}  \times 
\exp \left\{ -\frac{\lambda_k^W}{2} \sum_{i=1}^{M} \left(\sum_{j=1}^{K} w_{ij}^2\right) \right\}
\cdot u(w_{mk}) \\
&\propto \exp\left\{   -\frac{1}{2\sigma^2}  \sum_{j=1}^{N}(a_{mj} - \bw_m^\top\bz_j  )^2\right\}  \times   
\exp\left\{ -\frac{\lambda_k^W}{2}w_{mk}^2 \right\} 
\cdot u(w_{mk}) \\
&\propto \exp\left\{   -\frac{1}{2\sigma^2}  \sum_{j=1}^{N}
\left( w_{mk}^2z_{kj}^2 + 2w_{mk} z_{kj}\bigg(\sum_{i\neq k}^{K}w_{mi}z_{ij} - a_{mj}\bigg)  \right)
\right\}  \cdot \exp\left\{ -\frac{\lambda_k^W}{2}w_{mk}^2 \right\} \cdot u(w_{mk})\\
&\propto \exp\Bigg\{   
-
\underbrace{\left(\frac{\sum_{j=1}^{N} z_{kj}^2 + \textcolor{red}{\sigma^2\lambda_k^W} }{2\sigma^2}  \right)  }_{\textcolor{blue}{ 1/(2\widetilde{\sigma^2_{mk} }) }} 
w_{mk}^2 +  w_{mk}\underbrace{\left(  
\frac{1}{\sigma^2} \sum_{j=1}^{N} z_{kj}\bigg( a_{mj} - \sum_{i\neq k}^{K}w_{mi}z_{ij}\bigg)  \right)}_{\textcolor{blue}{\widetilde{\sigma_{mk}^{2}}^{-1} \widetilde{\mu_{mk}}}}
\Bigg\}  \cdot u(w_{mk})\\
&\propto   \normal(w_{mk} | \widetilde{\mu_{mk}}, \widetilde{\sigma_{mk}^{2}})\cdot u(w_{mk}) 
= \truncatednormal(w_{mk} | \widetilde{\mu_{mk}}, \widetilde{\sigma_{mk}^{2}}),
\end{aligned}
\end{equation}
where 
$$
\begin{aligned}
\widetilde{\sigma_{mk}^{2}}= \frac{\sigma^2}{\sum_{j=1}^{N} z_{kj}^2 +\textcolor{red}{\sigma^2\lambda_k^W}}, \gap \gap 
\widetilde{\mu_{mk}} = \left(
\frac{1}{\sigma^2} \sum_{j=1}^{N} z_{kj}\bigg( a_{mj} - \sum_{i\neq k}^{K}w_{mi}z_{ij}\bigg)  \right)\cdot \widetilde{\sigma_{mk}^{2}}
\end{aligned}
$$
are the posterior ``parent" variance of the normal distribution, and the posterior ``parent"  mean of the normal distribution respectively. 

\paragraph{Connection to GEE}
We observe that the ``parent" mean in the GL$_2^2$ model is larger than that in the GEE model (Eq.~\eqref{equation:gee_posterior_mean_append}) since it does not contain the negative term $-\lambda_{mk}^W$ in Eq.~\eqref{equation:gee_posterior_mean_append}. 
While the variance is smaller than that in the GEE model (Eq.~\eqref{equation:gee_posterior_variance_append}) such that the 
conditional density of GL$_2^2$ model is more clustered and it imposes a larger regularization.

\subsection{Gaussian Likelihood with $L_\infty$ Prior (GL$_\infty$) Model}\label{appendix:glinfity_regu_nmf}
\paragraph{Prior} We assume $\bW$ and $\bZ$  are independently distributed with parameter $\lambda_{k}^W$  and $\lambda_{k}^Z$ proportional to a exponential function: 
\begin{equation}\label{equation:gpinfty_prior_density_append}
\begin{aligned}
&p(\bW|\lambda_k^W) &\propto &
\left\{
\begin{aligned}
&\exp \left[  -{\lambda_k^W} \sum_{m=1}^{M} \mathop{\max}_{k} |w_{mk}| \right]
, &\gap &\text{if $w_{mk}\geq 0$ for all $m,k$ };
\\
&0, &\gap &\text{if otherwise};
\end{aligned}
\right.\\
\gap 
&p(\bZ|\lambda_k^Z) &\propto &
\left\{
\begin{aligned}
&\exp \left[  -\lambda_k^Z \sum_{n=1}^{N} \mathop{\max}_{k} |z_{kn}| \right]
, &\gap &\text{if $z_{kn}\geq 0$ for all $n,k$ };
\\
&0, &\gap &\text{if otherwise}.
\end{aligned}
\right.\\
\end{aligned}
\end{equation}
Note we remove the 2 in the denominator of $\lambda_k^W$ for consistency issue which we will see shortly in the form of the conditional density in Eq.~\eqref{equation:gpinfty_poster_wmk1_append}.
Again, the prior for the noise variance $\sigma^2$ is an inverse-gamma density with shape ${\alpha_\sigma}$ and scale ${\beta_\sigma}$,
$$
p(\sigma^2)= \inversegammadist(\sigma^2| \alpha_\sigma, \beta_\sigma) = \frac{{\beta_\sigma}^{\alpha_\sigma}}{\Gamma({\alpha_\sigma})} (\sigma^2)^{-\alpha_\sigma-1} \exp\left( -\frac{{\beta_\sigma}}{\sigma^2} \right).
$$
\paragraph{Posterior}
The conditional density of $\sigma^2$ is the same as that in the GEE model (Eq.~\eqref{equation:gee_posterior_sigma2_append}). 
Denote $\indicator(w_{mk})$ as the indicator whether $w_{mk}$ is the largest one among $k=1,2,\ldots, K$,
 the conditional density can be obtained by
\begin{equation}\label{equation:gpinfty_poster_wmk1_append}
\begin{aligned}
&\gap p(w_{mk}| \sigma^2, \bW_{-mk}, \bZ, \lambda_k^W,\cancel{\lambda_k^Z}, \bA)=p(w_{mk} | \sigma^2,   \bW_{-mk}, \bZ, \lambda_k^W, \bA) \\
&\propto p(\bA| \bW, \bZ, \sigma^2) \times p(\bW| \lambda_k^W)
=\prod_{i,j=1}^{M,N} \normal \left(a_{ij}| \bw_i^\top\bz_j, \sigma^2 \right)\times 
p(\bW| \lambda_k^W)
\cdot u(w_{mk}) \\
&\propto \exp\left\{   -\frac{1}{2\sigma^2}  \sum_{i,j=1}^{M,N}(a_{ij} - \bw_i^\top\bz_j  )^2\right\}  \times 
\exp \left\{ -{\lambda_k^W} \cdot\sum_{i=1}^{M} \mathop{\max}_{k} |w_{ij}| \right\}
\cdot u(w_{mk}) \\
&\propto \exp\left\{   -\frac{1}{2\sigma^2}  \sum_{j=1}^{N}(a_{mj} - \bw_m^\top\bz_j  )^2\right\}  \times   
\exp\left\{ -{\lambda_k^W}\cdot w_{mk} \right\} 
\cdot u(w_{mk}) \cdot \textcolor{red}{\indicator(w_{mk})}\\
&\propto \exp\left\{   -\frac{1}{2\sigma^2}  \sum_{j=1}^{N}
\left( w_{mk}^2z_{kj}^2 + 2w_{mk} z_{kj}\bigg(\sum_{i\neq k}^{K}w_{mi}z_{ij} - a_{mj}\bigg)  \right)
\right\}  \cdot \exp\left\{ - w_{mk}\cdot  \textcolor{red}{\lambda_k^W\cdot\indicator( w_{mk})}\right\}  \cdot u(w_{mk})\\
&\propto \exp\Bigg\{   
-
\underbrace{\left(\frac{\sum_{j=1}^{N} z_{kj}^2  }{2\sigma^2}  \right)  }_{\textcolor{blue}{ 1/(2\widetilde{\sigma^2_{mk} }) }} 
w_{mk}^2 +   w_{mk}\underbrace{\left(
-\textcolor{red}{\lambda_k^W\cdot \indicator(w_{mk})}+  
\frac{1}{\sigma^2} \sum_{j=1}^{N} z_{kj}\bigg( a_{mj} - \sum_{i\neq k}^{K}w_{mi}z_{ij}\bigg)  \right)}_{\textcolor{blue}{\widetilde{\sigma_{mk}^{2}}^{-1} \widetilde{\mu_{mk}}}}
\Bigg\}  \cdot u(w_{mk})\\
&\propto   \normal(w_{mk} | \widetilde{\mu_{mk}}, \widetilde{\sigma_{mk}^{2}})\cdot u(w_{mk}) 
= \truncatednormal(w_{mk} | \widetilde{\mu_{mk}}, \widetilde{\sigma_{mk}^{2}}),
\end{aligned}
\end{equation}
where 
$$
\begin{aligned}
\widetilde{\sigma_{mk}^{2}}= \frac{\sigma^2}{\sum_{j=1}^{N} z_{kj}^2 }, \gap \gap 
\widetilde{\mu_{mk}} = \left(
-\textcolor{red}{\lambda_k^W\cdot \indicator(w_{mk})}+
\frac{1}{\sigma^2} \sum_{j=1}^{N} z_{kj}\bigg( a_{mj} - \sum_{i\neq k}^{K}w_{mi}z_{ij}\bigg)  \right)\cdot \widetilde{\sigma_{mk}^{2}}
\end{aligned}
$$
are the posterior ``parent"  variance of the normal distribution, and the posterior ``parent"  mean of the normal distribution respectively. 

\paragraph{Connection to GEE}
We observe that the ``parent" variance in the GL$_\infty$ model is exactly the same as  that in the GEE model (Eq.~\eqref{equation:gee_posterior_variance_append}).
And when $\indicator(w_{mk})$ is satisfied, the ``parent" mean is the same as that in the GEE model as well (Eq.~\eqref{equation:gee_posterior_mean_append}).
However, when $w_{mk}$ is not the maximum value among $\{w_{m1}, w_{m2}, \ldots, w_{mK}\}$, the ``parent" mean is larger than that in the GEE model since the GL$_\infty$ model excludes this negative term.
The GL$_\infty$ model then has the interpretation that it has a \textit{sparsity constraint} when $w_{mk}$ is the maximum value; and it has a \textit{relatively loose constraint} when $w_{mk}$ is not the maximum value. Overall, the GL$_\infty$ favors a loose regularization compared with the GEE model.

\end{document}